\RequirePackage{fix-cm}
\documentclass[twocolumn]{svjour3}          
\smartqed  
\usepackage{graphicx}
\usepackage{amsmath}
\usepackage{amssymb}
\usepackage{xcolor}
\usepackage{physics}
\usepackage{qcircuit}
\usepackage{xcolor}
\usepackage{booktabs}
\usepackage[colorlinks=true,linkcolor=blue,allcolors=blue]{hyperref}%

\begin{document}

\title{Quantum Supervised Learning
}


\author{Antonio Macaluso 
}


\institute{A. Macaluso \at
Agents and Simulated Reality Department, German Research Center for Artificial Intelligence GmbH (DFKI), 66123 Saarbrucken, Germany \\
              \email{antonio.macaluso@dfki.de}           
}

\date{}

\maketitle

\begin{abstract}
Recent advancements in quantum computing have positioned it as a prospective solution for tackling intricate computational challenges, with supervised learning emerging as a particularly promising domain for its application. Despite this potential, the field of quantum machine learning is still in its early stages, and there persists a level of skepticism regarding a possible near-term quantum advantage.

This paper aims to provide a classical perspective on current quantum algorithms for supervised learning, effectively bridging traditional machine learning principles with advancements in quantum machine learning. Specifically, this study charts a research trajectory that diverges from the predominant focus of quantum machine learning literature, originating from the prerequisites of classical methodologies and elucidating the potential impact of quantum approaches. Through this exploration, our objective is to deepen the understanding of the convergence between classical and quantum methods, thereby laying the groundwork for future advancements in both domains and fostering the involvement of classical practitioners in the field of quantum machine learning.

\end{abstract}

\section{Introduction}
\label{sec:intro}

The adoption of machine learning has proliferated in recent years, largely due to the capacity for training large neural networks, which has significantly impacted various real-world applications. However, neural networks are computationally demanding, requiring specialized hardware and extensive datasets to learn meaningful representations \cite{lecun2015deep,krizhevsky2012imagenet,thompson2020computational}.
Conversely, traditional machine learning methodologies, such as linear regression \cite{chambers2017linear} and Support Vector Machines (SVMs) \cite{10.1214/009053607000000677,boser1992training} are grounded in assumptions about the data-generating process and do not typically achieve the state-of-the-art performance that neural networks can. 

With the advent of quantum computing, the question arises as to whether quantum approaches can enhance or surpass traditional machine learning methodologies. The inherent differences between quantum and classical computing frameworks carry significant implications for their application in supervised learning. While classical approaches typically rely on nonlinearity assumptions to uncover intricate patterns within datasets, with integration varying depending on the specific algorithm employed \cite{nair2010rectified,glorot2010understanding}, quantum computing leverages the principles of superposition, entanglement, and interference, enabling the execution of computational tasks that potentially exceed the capabilities of classical methods \cite{nielsen2010quantum}. However, quantum operations are governed by unitary and therefore linear transformations, which limit the direct translation of classical nonlinear models into quantum paradigms. Consequently, while quantum computing offers unparalleled computational efficiency, leveraging these advantages for supervised learning needs a reassessment to align with quantum principles.

The continuous advancement of quantum technology has motivated researchers to develop innovative strategies for applying quantum computing to traditional problems such as regression \cite{reddy2021hybrid,wang2023variational}, classification \cite{schuld2020circuit}, and function approximation \cite{10.1007/978-3-031-36030-5_14,macaluso2020quantum}. However, Quantum Machine Learning (QML) models, still in their embryonic stages, have yet to offer a feasible alternative to the entrenched classical methods. This is partly because current classical methodologies are largely driven by extensive experiments, while conducting such experiments with the existing quantum hardware is exceedingly difficult.


In this paper, we examine the QML literature from a traditional standpoint, establishing a connection between classical and quantum supervised learning techniques.
The objective here is 
to explicate what quantum computing entails for the field of machine learning and to identify potential synergies between the two domains. 
Specifically, we begin by presenting an overview of the problem of supervised learning, clarifying the crucial distinction between parametric and non-parametric methods for adaptation in quantum settings (Section \ref{sec:supervised}). Consequently, we outline the QML landscape, highlighting how varying assumptions about the capabilities of the quantum computer at hand lead to distinct QML strategies and the potential enhancements quantum computing may introduce (Section \ref{sec:QML}). In Section \ref{sec:current}, we delve into the latest approaches and trends in QML research, emphasizing their potential impact on the field of classical machine learning. 
Following this, we identify several challenges that arise from the requirements of contemporary classical methods and discuss how quantum techniques could address these issues (outlined in Section \ref{sec:primising}). The paper concludes with a synthesis of our contributions in the concluding section (Section \ref{sec:conclusion}).

\section{Supervised Learning: A Methodological Perspective}
\label{sec:supervised}


Supervised methods \cite{10.5555/1162264,vapnik1991principles} aim to learn an unknown target function $f: \mathcal{X} \rightarrow \mathcal{Y}$, where $\mathcal{X}$ and $\mathcal{Y}$ are respectively the sets of features and the target variable. Each learning algorithm is associated with a hypothesis class $\mathcal{H}$ comprising functions $h: \mathcal{X} \rightarrow \mathcal{Y}$.
The set $\mathcal{X}$ serves as the domain for $\mathcal{H}$, which consists of objects with bounded computational power. For instance, $\mathcal{H}$ might include all neural networks with specific and possibly limited depths and number of nodes, or all decision trees of at most a certain depth.
It is assumed that each function $h \in \mathcal{H}$ possesses a succinct description, and is feasible to evaluate a given $h$ on a given $x \in \mathcal{X}$.
The goal of the learner\footnote{The definition refers to classification, but it can be easily generalized to regression.} is to minimize the \textit{generalization error} of $h$ with respect to   $f$:
\begin{align}\label{eq:generalization}
    err(h, f, \mathcal{D}) = \underset{x \sim \mathcal{D}}{Pr}\left[h(x) \neq f(x)\right],
\end{align}
where $\mathcal{D}$ is the probability distribution over $\mathcal{X}$. Generalization error (Eq. \eqref{eq:generalization}) is minimized by the learner over the class $\mathcal{H}$. However, since $\mathcal{D}$ and $f$ are in general unknown, the generalization error is not directly available. 

In practice, the learner has access to a limited training set $S=\{(x^{(i)}, y^{(i)})\}_{i=1, \dots, N}$, where the points $x^{(i)}$ are presumed \textit{independent and identically distributed}, and generated according to the unknown distribution $\mathcal{D}$ on $\mathcal{X}$. The target variable can be computed as $f(x^{(i)})=y^{(i)}$ and $f$ represents the function to learn. 
A useful metric of error is the training error which is computed according to the instances in $S$:
\begin{align}
    \hat{err}(h, f, \mathcal{D}) = \underset{x,y \in S}{Pr}\left[h(x) \neq f(x)\right].
\end{align}
%
This is also known as empirical risk \cite{vapnik1991principles}. Since the training set is a snapshot of the $\mathcal{D}$, it makes sense to search for solutions that work well on the training data. This learning paradigm is called \textit{Empirical Risk Minimisation} \cite{vapnik1991principles} and aims to find a predictor $h$ that minimizes $ \hat{err}(h, f, \mathcal{D})$, in the hope that it will be able to generalize over the unknown $\mathcal{D}$. 
Therefore, in essence, machine learning methods refer to a set of approaches for estimating $f$ that can be characterized as either parametric or non-parametric \cite{russell2010artificial}.

\paragraph{Parametric Methods}

This class of methods employs a model-based strategy that unfolds in two primary phases. Initially, a specific functional form for $f$ is posited \emph{(Model Specification)}. A common starting point is to assume a linear relationship with respect to $x$, articulated as:
\begin{align}\label{eq:lin_reg}
f(x) = \beta_0 + \beta_1 x_1 + \beta_2 x_2 + \dots + \beta_p x_p.
\end{align}

This presumption of linearity greatly streamlines the model estimation process. Rather than deducing a completely arbitrary function $f(x)$ in a $p$-dimensional space, the task is reduced to determining the $p + 1$ coefficients $\beta_0, \beta_1, \dots , \beta_p$. 

The next phase involves fitting the model with training data. For the linear framework described above, the goal is to ascertain the values of $\beta_0, \beta_1, \dots , \beta_p$. By making specific assumptions about $f$ and utilizing the residual sum of squares as the loss function, the problem is transformed into a convex optimization task \cite{kutner2005applied,hastie01statisticallearning}. This allows for the derivation of the optimal $\beta$ coefficients for a given dataset through least squares, yielding a closed-form solution.
In matrix form, if $X$ denotes the $N \times (p+1)$ matrix with each row representing an input vector (with a $1$ in the first position), and $y$ denotes the $N$-vector of outputs in the training set, the residual sum of squares (RSS) can be rewritten as:
\begin{align}
RSS(\beta)= (y-X\beta)^T(y-X\beta).
\end{align}
This expression represents a quadratic form in the $p+1$ parameters. By differentiating with respect to $\beta$ and assuming that $X$ has full column rank and hence $X^T X$ is positive definite, the estimate of $\beta$ that minimizes the function $RSS(\beta)$ is given by:
\begin{align}\label{eq: beta linear regression system}
\hat{\beta}=(X^TX)^{-1}X^T y.
\end{align}

Therefore, to fit a linear model following the minimization of $RSS(\beta)$, it is necessary to solve the linear system expressed in Eq. (\ref{eq: beta linear regression system}).
It is important to note that the variables $X$ are not necessarily the observed features. For instance, the input features can be augmented with basis expansions \cite{hastie2009basis} to introduce non-linearity through spline functions \cite{de1978practical}. Splines divide the sample data into sub-intervals delimited by breakpoints. A fixed-degree polynomial is then fitted within each segment, resulting in a piecewise polynomial regression.

While assuming a parametric form for $f$ simplifies the estimation process, selecting a model that deviates significantly from the true $f$ can lead to inaccurate estimates. This issue can be mitigated by employing more flexible models capable of accommodating various potential functional forms for $f$. However, employing more complex models often entails estimating a greater number of parameters, potentially leading to overfitting, where the model excessively conforms to the noise present in the data \cite{hastie01statisticallearning}.


\paragraph{Non-Parametric Methods}

Non-parametric methods eschew explicit assumptions regarding the functional form of $f$, opting instead to seek an estimate of $f$ that closely approximates the data points while maintaining a balance between smoothness and fidelity. This characteristic endows non-parametric approaches with a significant advantage over parametric methods: by sidestepping the need to specify a particular functional form for $f$, they possess the capability to accurately capture a broader spectrum of potential shapes for $f$. 
However, 
non-parametric techniques suffer from a notable drawback: due to their failure to reduce the task of estimating $f$ to a limited number of parameters, they necessitate a substantially larger number of observations and imply more complex optimization approaches. 

Prominent examples of this paradigm are Support Vector Machines (SVMs) \cite{boser1992training} and Neural Networks \cite{lecun2015deep}. Both can be categorized as non-parametric models due to their flexibility in capturing complex patterns without relying on predefined functional forms. However, there exists a fundamental difference in their underlying assumptions about $f$. SVMs, while non-parametric, make assumptions about the data's linear separability in a predetermined feature space $\phi(\cdot)$, leading to a convex optimization problem. 
Mathematically, for a binary classification problem, the decision function of SVMs can be represented as \cite{boser1992training}:
\begin{align}\label{eq:classical_kernel}
    f(x) = \text{sign}\left( \sum_{i=1}^{N} \alpha_i y^{(i)} k(x^{(i)}, x) + b \right)    
\end{align}

where $x$ represents the input feature vector, $\alpha_i$ are the Lagrange multipliers associated with the support vectors corresponding to the target values $y_i$, $b$ is the bias term and $k(x^{(i)}, x) = \langle \phi(x^{(i)}), \phi(x) \rangle$ is the kernel function that computes the similarity, in terms of inner product, between $x$ and $x^{(i)}$ in the feature space $\phi(\cdot)$.

SVMs work by implicitly mapping input data into a high-dimensional feature space where they can be linearly separated but without explicitly computing the transformation itself. The choice of kernel function determines the shape of the decision boundary. According to this interpretation, SVMs can be understood and analyzed with respect to the reproducing kernel Hilbert space (RKHS) \cite{smola1998learning} that is a mathematical framework characterized by being a vector space of functions endowed with an inner product operation.

A specific case of SVMs consists of assuming the least-squares criterion for the error terms and transforming the inequality constraints into equality constraints. This approach, known as Least-squares SVMs (LS-SVM) \cite{boser1992training}, leads to an optimization problem that aligns more closely with parametric model settings. 

\begin{figure*}[ht]
    \centering
    \hspace*{-1em}
    \includegraphics[scale=.3]{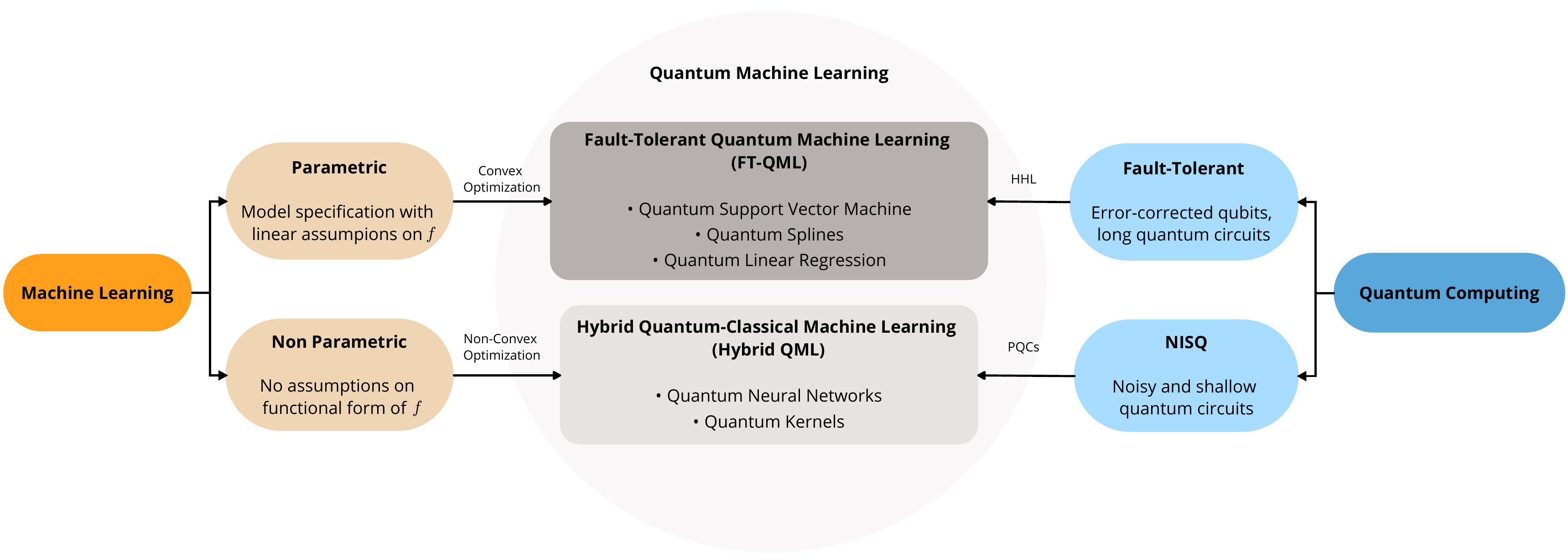}
        \caption{
        Visual representation of the field of Quantum Machine Learning as the intersection between different paradigms in machine learning (parametric and non-parametric) and quantum computation capabilities (Fault-tolerant and NISQ). Fault-Tolerant Quantum Machine Learning (FT-QML) involves quantum algorithms like Quantum Support Vector Machines \cite{rebentrost2014quantum} (least-square formulation \cite{ye2007svm}), Quantum Splines \cite{macaluso2020quantum}, and Quantum Linear Regression \cite{PhysRevA.96.012335}, which require error-corrected qubits and the capability to run arbitrarily long quantum circuits. Conversely, Hybrid QML incorporates Quantum Neural Networks \cite{macaluso2020variational} and Quantum Kernels \cite{mengoni2019kernel}, utilizing NISQ devices characterized by noisy and shallow quantum circuits. These approaches integrate with traditional machine learning methods. In particular, parametric models, which assume linearity in the underlying function $f$ and rely on convex optimization procedures, can benefit from the HHL algorithm \cite{PhysRevLett.103.150502} in the FT-QML setting. On the other hand, non-parametric models, which do not impose such assumptions, can be enhanced by utilizing parametrized quantum circuits (PQCs) to introduce a new class of hypotheses in the Hybrid QML settings.  }
\label{schema QML}
\end{figure*}

Differently from SVMs, neural networks \cite{lecun2015deep} do not impose any explicit assumptions on the data's structure, providing more flexibility in modeling complex relationships without necessarily requiring linear separability in a predefined feature space. This flexibility comes at the cost of increased computational complexity, a non-convex optimization landscape, and the need for larger datasets to effectively learn the underlying patterns.
Over the past decade, the proliferation of data availability combined with the adoption of GPUs has accelerated the development of large neural networks. This empirically driven methodology presents a challenge to the traditional reliance on theoretical constructs such as VC dimension \cite{vapnik1994measuring} or effective dimension \cite{berezniuk2020scale} to gauge a model's generalization capacity \cite{zhang2021understanding}. This is because classical statistical learning theory does not always correspond with the overparameterized regime prevalent in deep learning \cite{allen2019learning}, which frequently surpasses other methods. These developments prompt questions about the suitability of these theoretical frameworks in accurately assessing a model's true effectiveness \cite{zhang2021understanding,zhang2017understanding,NIPS2017_10ce03a1}. 


\section{Quantum Machine Learning Landscape}
\label{sec:QML}

The field of Quantum Machine Learning (QML) sits at the intersection of machine learning and quantum computation, aiming to leverage quantum computational tools for innovative solutions to conventional machine learning problems. Within QML, two primary methodologies outline distinct approaches based on the nature of the underlying machine learning techniques, dictating the computational tasks assigned to quantum computers. These approaches share similarities with the dichotomy between parametric and non-parametric classical models discussed in Section \ref{sec:supervised}.

The fault-tolerant approach \cite{Schuld2021} aims to develop quantum algorithms that can significantly accelerate machine learning tasks, with the ambition of achieving demonstrable exponential speed-up. These algorithms strive to emulate the outcomes of classical processes \textit{faster} in terms of computational complexity theory. 
Nevertheless, this approach usually requires a fully error-corrected quantum computer and the proposed algorithms tend to align with the long-term prospects of quantum computing rather than immediate practical applications. Conversely, the near-term trajectory of QML \cite{gujju2023quantum,wang2024quantum} focuses on leveraging capabilities of Noisy intermediate-scale quantum computers (NISQ) to develop novel models and training algorithms \cite{cheng2023noisy,preskill2018quantum}. This paradigm arises from the need to work with a limited number of non-error-corrected qubits and shallow quantum circuits.

A visual description of QML from both machine learning and quantum computing perspectives is shown in Figure \ref{schema QML}.

\begin{figure*}
    \centering
    \includegraphics[scale=.63]
    {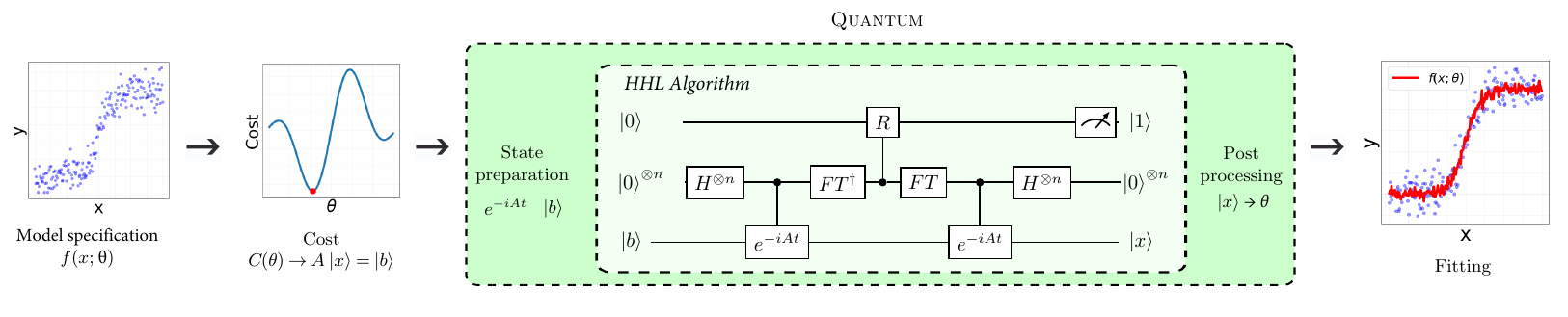}
        \caption{Schema for implementing a fault-tolerant quantum machine learning approach based on the HHL algorithm. The first step involves the model specification, where a preemptive assumption is made about the linear relationship between the target variable of interest and the basis expansion of the input feature \(x\), such as in the case of spline functions. The second step is to formulate a linear system of equations of the form \(A\ket{x} = \ket{b}\). It is important to note that in this context, the quantum state \(\ket{x}\) does not correspond to the input \(x\) but instead contains information about the classical set of parameters \(\theta\). Once the linear system is established, the task of finding its solution is assigned to the HHL algorithm, which primarily comprises three sub-steps: (i) state preparation, involving the quantum gate \(e^{iAt}\) and quantum state \(\ket{b}\); (ii) execution of the HHL quantum circuit; and (iii) post-processing of \(\ket{x}\) to classically extract the relevant information for estimating \(\theta\), (the swap-test in the case of quantum splines). Finally, the parameters obtained through HHL are classically used to estimate the function of interest.
}
    \label{fig:FT}
\end{figure*}

\subsection{Fault-Tolerant Quantum Machine Learning}

Fault-tolerant Quantum Machine Learning (FT-QML) endeavors to enhance classical computational routines by leveraging fault-tolerant quantum devices.
In particular, a suite of quantum algorithms known as \textit{quantum BLAS} (Basic Linear Algebra Subroutines) \cite{ciliberto2018quantum,biamonte2017quantum} aims to facilitate the execution of fundamental operations such as matrix multiplication, inversion, and singular value decomposition at a computational pace surpassing that of classical counterparts. Within the FT-QML paradigm, quantum algorithms may manifest logarithmic runtime complexity relative to the input dimension \( N \) and \( p \), contingent upon the complexity achieved by the input encoding. 

This methodology is particularly advantageous for parametric models, wherein the optimization problem simplifies conducting linear algebraic operations to align a given model with the data. However, this approach poses several challenges, with the primary obstacle being the necessity for a fault-tolerant quantum computer equipped with a substantial number of error-corrected qubits to enable execution \cite{paler2015introduction}.

Importantly, the primary focus of FT-QML is not to augment the learning capabilities of models but rather to expedite computational speed in terms of worst-case theoretical complexity relative to existing classical algorithms. This is achieved by delegating to a quantum computer a single heavy computational task that potentially requires a single call to a quantum computer (as described in Figure \ref{fig:FT}). 


\subsubsection{Quantum Linear Systems of Equations}\label{subsec: HHL}

HHL \cite{PhysRevLett.103.150502} constitutes a quantum algorithm designed to approximate the preparation of a quantum superposition denoted as $\ket{x}$, where $x$ denotes the solution to a linear system $Ax = b$, under the assumption of efficient preparation of the state $\ket{b}$ and the application of the unitary transformation $e^{-iAt}$. Its time complexity scales approximately as $\mathcal{O}(s^2\kappa^2\log(N)/\epsilon)$, where $N$ represents the system size, $\kappa$ denotes the system's condition number, $s$ indicates its sparsity and $\epsilon$ signifies the desired error \cite{wiebe2012quantum}.
Mathematically, the objective of the HHL algorithm, given an $N \times N$ Hermitian matrix $A$ and a unit vector $b$, is to determine the solution vector $x$ that satisfies the equation:
\begin{equation}\label{HHL: system of equations}
A \ket{x} = \ket{b},
\end{equation}
where vectors $b$ and $x$ are translated into the amplitudes of the corresponding quantum registers $\ket{b}$ and $\ket{x}$. 


Despite its exponential advantage concerning system size over state-of-the-art classical alternatives 
\cite{ciliberto2018quantum,burgisser2013algebraic,coppersmith1987matrix,strassen1969gaussian,shewchuk1994introduction} (a comprehensive description is shown in Table \ref{tab:HHL_efficiency}), several caveats constrain its applicability to practical problems \cite{aaronson2015read}. Firstly, it necessitates the matrix $A$ to be sparse due to the polynomial dependency on the level of sparsity $s$. Secondly, data loading of the vector $b$ into quantum superposition must be achieved efficiently to preserve the computational advantage for inverting $A$. Thirdly, the output is encoded in a quantum state $\ket{x}$, where the coefficients' approximation may be prohibitive for large linear systems, as all entries of $x$ are in superposition. Fourthly, the condition number and sparsity must scale at most sub-linearly with $N$. Finally, extracting all $N$ amplitudes of an $N$-dimensional quantum state requires a number of measurements at least proportional to $N$. Thus, if the goal is a complete reconstruction of a solution $x$, the quantum algorithm cannot possess a significant advantage over classical methods \cite{aaronson2015read}. 
Nevertheless, the exponential advantage in the size of the matrix 
$A$ is something that is not achievable classically, paving the way for the development of quantum-supported algorithms that meet constraints on other parameters and deliver an exponential speedup when adopted for optimized parametric models in machine learning.
In the following, we describe two methodologies that utilize the HHL algorithm to achieve a computational advantage over classical methods, while also introducing a method for incorporating non-linearity using a quantum algorithm.

\begin{table*}[t]
\begin{center}
\begin{tabular}{c c c c c}
\toprule
\parbox{2cm}{\textbf{Gauss \\ Jordan}}  & \textbf{Strassen} & \textbf{Coppersmith} & \parbox{2cm}{\textbf{Conjugate \\ Gradient}} & \textbf{HHL} \\[2mm]
\toprule
$\mathcal{O}(N^3)$ & $\mathcal{O}(N^{2.8})$ & $\mathcal{O}(N^{2.37})$ & $\mathcal{O}(sN\sqrt{\kappa}/\text{log}(\epsilon))$ & $\mathcal{O}(s^2\kappa^2\text{log}(N)/\epsilon)$ \\
\bottomrule
\end{tabular}
{\caption{Comparison of algorithms computational costs for solving a linear system of equations. In general, matrix inversion can be accomplished in polynomial time on classical devices \cite{burgisser2013algebraic,coppersmith1987matrix,strassen1969gaussian} of size $N$. However, under several favorable assumptions on $\kappa$ (condition number) and $s$ (sparsity), it is possible to reduce computational costs using \textit{Conjugate Gradient} \cite{ciliberto2018quantum,shewchuk1994introduction}.}
\label{tab:HHL_efficiency}}
\end{center}
\end{table*}

\subsubsection{Least-square Support Vector Machines}


In the context of least-squares SVMs (LS-SVMs) \cite{ye2007svm}, the optimization problem is reformulated into a least squares problem, simplifying the solution process but necessitating the inversion of the kernel Gram matrix ($K$). This matrix encapsulates pairwise similarities between training samples computed by the kernel function. LS-SVMs require solving the following linear system of equations: 
\begin{align}\label{eq:ls-svm}
    K \begin{pmatrix} w_0\\ \gamma \end{pmatrix}  = \begin{pmatrix} 0 \\ y \end{pmatrix},
\end{align}
where $\gamma$ denotes the Lagrange multipliers and $w_0$ represents the bias term. 
Notice that the generation of the matrix 
$K$ implicitly encodes the concept of a kernel, utilizing a specific feature map that enables the mapping of data from the input feature space to a higher-dimensional feature space where data points become linearly separable (the foundational assumption of SVMs), implicitly introducing non-linearity.

Thus, training LS-SVMs consists of solving a convex quadratic optimization problem (Eq. \eqref{eq:ls-svm}) and is typically addressed using classical algorithms, with computational complexity polynomial in terms of the number of features ($p$) and training samples ($N$), and inversely proportional to the desired accuracy ($\epsilon$), typically expressed as $\mathcal{O}(\log(\epsilon^{-1})\text{poly}(p, N))$ \cite{abdiansah2015time}. The adoption of HHL has been proposed to introduce a quantum SVM protocol with time complexity of $\mathcal{O}(\log(pN))$ for both training and testing, achieved by leveraging principles of quantum mechanics \cite{rebentrost2014quantum}.


Despite the theoretical advantage, a fully operational quantum SVM applicable to real-world datasets remains unrealized. 
Nevertheless, the potential for quantum resources to significantly reduce the computational burden of training SVMs holds promise for expanding their applicability in various machine learning tasks, as long as a fault tolerant quantum computer for executing the HHL is available.

\subsubsection{Quantum Splines for Non-Linear Approximation}

Spline functions \cite{deboor} are methods for fitting non-linear functions through the basis expansion of input features. Specifically, spline estimation involves segmenting data into intervals defined by knots, with a polynomial fitted within each interval, thus forming a piecewise polynomial regression. However, the computational efficiency of spline methods based on truncated basis functions is often suboptimal, which leads to a preference for the B-splines approach in practical applications \cite{de1972calculating}. 

B-splines generate a design matrix characterized by a consistent level of sparsity, which is determined by the degree of the local polynomials.
For a set of knots $\{\xi_1, \xi_2, \ldots, \xi_T\}$, a polynomial is fitted within each interval $[\xi_k, \xi_{k+1}]$ for $k = 1, \ldots, T-1$, without imposing continuity of derivatives across knots. The model can be represented by a linear system:
\begin{equation}\label{eq:B-splines}
\tilde{y} = S\beta \rightarrow 
\begin{pmatrix}
\tilde{y}_1 \\
\tilde{y}_2 \\
\vdots \\ 
\tilde{y}_{K}\\
\end{pmatrix}
=
\begin{pmatrix}
S_1 & 0 & \cdots & 0\\
0 & S_2 & \cdots & 0\\
\vdots & \vdots    & \ddots & \vdots\\ 
0 & 0 & \cdots & S_{K}\\
\end{pmatrix} 
\begin{pmatrix}
\beta_1 \\
\beta_2 \\
\vdots \\ 
\beta_{K}\\
\end{pmatrix},
\end{equation}
where $\tilde{y}_k$ represents function evaluations between $\xi_k$ and $\xi_{k+1}$, $\beta_k$ are spline coefficients, and $S$ is a block diagonal matrix with each block $S_k$ corresponding to basis functions in the $k$-th interval. Solving this linear system yields spline coefficients approximating the non-linear functions captured in $\tilde{y}$.
Therefore, the non-linear approximations through spline functions fall within the parametric framework of linear regression, enabling the estimation of the optimal parameters $\beta$ by solving a linear system of equations (as shown in Eq. \eqref{eq: beta linear regression system}).

Quantum Splines (QSplines) \cite{macaluso2020quantum} extend B-splines to quantum computing, aiming to leverage quantum algorithms for function approximation. In order to experimentally solve the linear system of equations (Eq. \eqref{eq:B-splines}) using HHL,  QSplines formulation exploits the fact that the inverse of a block diagonal matrix remains block diagonal, with corresponding inverse matrices in each block. Therefore, QSplines implementation solves $K$ $2 \times 2$ quantum linear systems $S_k \ket{\beta_k} = \ket{\Tilde{y_k}}$ instead of a single one for the entire function. This approach overcomes the practical limitations of available quantum simulators, enabling the calculation of spline coefficients through quantum simulations.

The computation of a QSpline involves three steps. First, HHL computes spline coefficients for the $k$-th interval:
\begin{equation}
S_k \ket{\beta_k} = \ket{\Tilde{y_k}} \xrightarrow{HHL} \ket{\beta_k} \simeq S_k^{-1} \ket{\Tilde{y_k}}.
\end{equation}
Second, $\ket{\beta_k}$ interacts with the quantum state encoding input in the $k$-th interval via quantum interference. The scalar product between $\ket{\beta_k}$ and $\ket{x_{i,k}}$ is computed using the swap-test \cite{PhysRevLett.87.167902}:
\begin{align}
    \ket{\beta_k} \ket{x_{i,k}} \ket{0}  \xrightarrow{swap-test} \ket{e_1} \ket{e_2} \ket{f_{i,k}}.
\end{align}
At this point, amplitudes of quantum state $\ket{f_{i,k}}$ embed the estimate of the activation function evaluated in $x_{i,k}$. Third, $\ket{f_{i,k}}$ is measured to obtain probability of state $\ket{0}$, depending on dot product between $\beta_k$ and $x_{i,k}$:
\begin{align}
    \ket{f_{i,k}} = \sqrt{p_0}\ket{0} + \sqrt{p_1}\ket{1},
\end{align}
where: 
\begin{align} \label{eq:p_0}
    p_0 = \frac{1}{2}+\frac{|\langle \beta_k| x_{i,k} \rangle|^2}{2} = \frac{1}{2}+\frac{|f_{i,k}|^2}{2} .
\end{align}
Finally, the non-linear function estimate in correspondence of $x_{i,k}$ is retrieved by back-transforming Eq. \eqref{eq:p_0} to get $f_{i,k}$.
Notice that, the estimates are intrinsically bounded in the interval $[0,1]$ since they are encoded as the amplitude of a quantum state. 

\begin{figure*}
    \centering
    \includegraphics[scale=1]
    {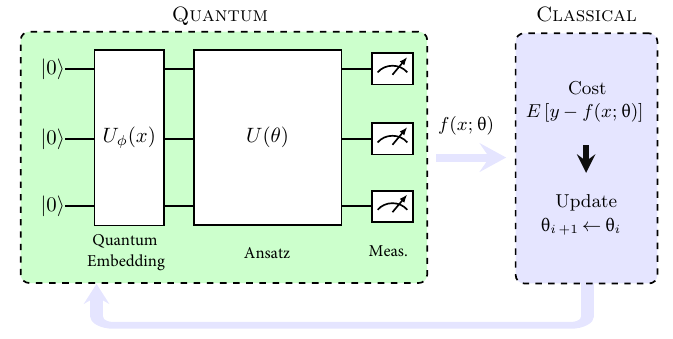}
        \caption{Scheme of a hybrid quantum-classical algorithm for supervised learning (adapted from \cite{macaluso2022variational}). The quantum variational circuit is depicted in green, while the classical component is represented in blue.}
    \label{fig:hybrid}
\end{figure*}

The development of QSplines holds significance in several key areas. Firstly, the size of the linear system generated by spline formulations escalates substantially, being directly proportional to both the number of training points (assumed to equal the number of knots) and the dataset's feature count. This scalability issue renders traditional spline methods impractical for large-scale data problems, where the matrix size becomes a critical factor in computational efficiency.

Secondly, when considering B-splines, the resulting linear system is characteristically sparse, which aligns well with the HHL algorithm's capability to exploit such sparsity for efficiency gains, particularly in terms of the parameter 
$s$. Assuming the practical application of the HHL algorithm on quantum hardware, the computational challenges associated with spline functions could be significantly reduced. This advancement opens the door to introducing non-linearity by means of HHL, thus overcoming the constraint about the unitarity of quantum computation.

\subsection{Hybrid Quantum-Classical Machine Learning}

The construction of full-scale, error-corrected quantum devices still presents numerous technical challenges, and implementing algorithms such as the HHL remains unfeasible given the current state of quantum technology. At the same time, significant progress has been made in the development of small-scale quantum computers, thus giving rise to the so-called Noisy Intermediate-Scale Quantum (NISQ) era \cite{bharti2022noisy,cheng2023noisy}.
Therefore, many researchers are currently focusing on algorithms for NISQ machines that may have an immediate impact on real-world applications.
Such machines, however, are still not sufficiently powerful to be a credible alternative to the classical ones.
For this reason, \textit{hybrid computation} was proposed to exploit near-term devices to benefit from the performance boost expected from quantum technologies. 
Quantum variational algorithms \cite{PhysRevA.92.042303,moll2018quantum} represent the most promising attempt in this direction, and they are designed to tackle optimization problems using both classical and quantum resources.
The latter component is referred to as variational circuit. 

As discussed in Section \ref{sec:supervised}, supervised learning problems typically involve fitting a parameterized function to a training dataset. In the context of Hybrid QML, a hybrid computational approach employs Parametrized Quantum Circuits (PQCs) to define a class of hypothesis functions. These functions are used to estimate a target variable of interest in a supervised learning setting, to achieve representational capabilities beyond those of classical methods. The general hybrid approach is illustrated in Figure \ref{fig:hybrid}. The data, $x$, are initially pre-processed on a classical device to determine a normalize input quantum state.
Firstly, we consider a feature encoding  unitary $U_\phi: \mathcal{X} \rightarrow \mathcal{F}$ that maps the input vector $x \in \mathcal{X}$ to a $n$-qubit quantum state $\ket{\phi(x)} = U_\phi(x)\ket{0}$ in the Hilbert space $\mathcal{F}$ of $2^n \times 2^n$ Hermitian operators. 
 Then, the quantum hardware prepares a quantum state $\ket{\phi(x)}$  and computes $U(\theta)$ with randomly initialized parameters $\theta$. After multiple executions of $U(\theta)$ (ansatz), the classical component post-processes the measurements and generates a prediction $f(x; \theta)$. Finally, the parameters are updated, and the whole cycle is run multiple times in a closed loop between the classical and quantum hardware. 
The strength of this approach is the possibility of learning gate parameters enables the adaptation of the architecture for different use cases.

The Hybrid QML approach shares significant similarities with the training of classical neural networks, notably in their reliance on parameterized models, gradient-based optimization techniques, and structured layers for approximating complex functions. This common ground has paved the way for the development of Quantum Neural Networks (QNNs), which are essentially the use PQCs for machine learning applications. The primary distinction between QNNs based on hybrid computation and classical neural networks lies in the execution of function calls to \( f(x; \theta) \): in the former, a PQC is employed, while in the latter, a classical neural model is used. This subtle yet fundamental difference aims to leverage the unique capabilities of quantum computing, potentially offering advantages in processing efficiency and learning capabilities for specific tasks.

It is important to note the distinction between FT-QML and Hybrid QML based on PQCs. While FT-QML primarily leverages the quantum component to expedite the training process, the use of PQCs in QML seeks to identify problem classes that are intractable for classical approaches from a learning perspective. 
Furthermore, another significant difference lies in the frequency of calls to a quantum computer during the training procedure. In the case of FT-QML, the (error-corrected) quantum computer is usually invoked only once to solve the linear system of equations encoding the training data. In the hybrid approach, the quantum computer is tasked with making (at least) one call to the function 
$f(x;\theta)$ for each iteration of the training process.

\subsubsection{Quantum Models as Kernel Methods}

In the implementation of QNNs, the initial step involves representing data in the form of a quantum state, a process for which various methodologies have been proposed \cite{weigold2021encoding}. Prior research has established a link between quantum computational models and kernel methods from classical machine learning \cite{gil2023expressivity,schuld2021supervisedqml}. 
For example, in the context of SVMs, the objective is to identify the optimal hyperplane that can separate data points of differing classes within a high-dimensional feature space $\phi(\cdot)$. The kernel trick employed by SVMs facilitates the implicit mapping of input features to this higher-dimensional space, computing only the inner products between the transformed feature vectors explicitly.

Conversely, when encoding data into a quantum state, the mapping of the input $x$ is executed explicitly through the definition of a series of quantum gates $U_\phi(x)$. 
This explicit approach contrasts with the implicit nature of the kernel trick in classical SVMs, highlighting a fundamental difference in how data transformation is achieved in quantum versus classical machine learning paradigms.
Still, this similarity allows studying quantum kernels using the RKHS framework \cite{smola1998learning}, making a strong methodological connection between the two approaches. 

In concrete terms, quantum kernels offer two distinct methodologies for enhancing machine learning tasks \cite{Schuld2021}. The first strategy involves utilizing a quantum device to estimate the inner products of quantum states $k(x^{(i)}, x) = \langle \phi(x^{(i)}), \phi(x) \rangle$ in Eq. \eqref{eq:classical_kernel}, thereby facilitating the computation of a kernel that may be classically intractable. This quantum-derived kernel undergoes subsequent classical post-processing and integrates into traditional machine learning frameworks to predict the target variable of interest. It is crucial to note that in this scenario, the quantum component primarily augments the kernel computation, which is subsequently applied within conventional kernel-based methods, as illustrated in Equation \ref{eq:classical_kernel}.

Conversely, the second approach employs a PQC as an explicit classifier within the Hilbert space. Here, the PQC operates as a linear model, directly classifying data points by leveraging the high-dimensional space of quantum states. This method stands in contrast to the kernel-based approach by providing a more direct utilization of quantum computational resources for supervised tasks.

In the outlined quantum methodologies, both can be conceptualized as families of hypotheses, wherein the objective is to identify a function within the RKHS framework \cite{smola1998learning} that exhibits optimal learning performance. Notably, beyond the aforementioned quantum approaches, the technique of data re-uploading \cite{perez2020data} has recently gained recognition for its efficacy in accommodating inputs of varying sizes in relation to the number of qubits. This method serves as a universal approximator and still is theoretically aligned with the RKHS framework for quantum kernels \cite{jerbi2023quantum}. 

Therefore, when interpreting QML models as kernel-based methodologies, the objective centers on delineating kernel functions that, while challenging to compute via classical means, can be adeptly realized through the utilization of shallow PQCs. While empirical validation for such kernels has been demonstrated in the case of contrived, problem-specific scenarios \cite{havlivcek2019supervised}, a definitive demonstration of quantum advantage through this modality on real applications remains elusive.

\subsubsection{Classical-Inspired Quantum Models}\label{sec:classical-inspired}

An alternative strategy in building hybrid QML models based on PQCs involves the development of parameterized ansätze inspired by classical machine learning techniques. A prime example of this approach are the Quantum Convolutional Neural Networks (QCNNs) \cite{cong2019quantum}, which draw inspiration from its classical counterpart.
In a QCNN, an unknown quantum state, denoted by $\rho_{\text{in}}$, is fed as input into the circuit. The first stage involves a quantum convolution layer that applies a quasi-local unitary operation, $U_i$, across the input in a translationally invariant manner, ensuring a finite depth of the network. Subsequent pooling is achieved by measuring a subset of qubits, with the outcomes of these measurements guiding the application of unitary rotations on adjacent qubits. This mechanism aim to introduce nonlinear dynamics into the QCNN by reducing the system's degrees of freedom as it progresses through the layers.
The process of convolution and pooling is iterated until the quantum system is reduced to a sufficiently small size. At this juncture, a fully connected layer, characterized by a unitary transformation is applied to the remaining qubits. The QCNN's output is then derived from measurements performed on a specific set of output qubits.

Mirroring the design principles of classical Convolutional Neural Networks (CNNs), the architecture of a QCNN, including the number and arrangement of convolution and pooling layers, is predefined. The unitary transformations within the network are the variables optimized during the learning phase. As a result, a QCNN configured to classify $N$-qubit input states is distinguished by a parameter complexity of $O(\log(N))$, indicating a scalable and efficient framework for quantum machine learning \cite{cong2019quantum}.

Notably, QCNNs have demonstrated good generalization performance even when trained on a limited set of training data for supervised classification of quantum states \cite{caro2022generalization}. Although the application domain is primarily relevant to physics, the ability to train a model with a significantly small amount of training data is a crucial aspect of classical machine learning.

An alternative approach to QCNNs involves constructing a quantum ansatz capable of generating a quantum state that mirrors the output of classical single-layer neural networks \cite{macaluso2020variational,macaluso2022variational,macaluso2023maqa}. Given the universal approximation capability of these networks \cite{hornik1989multilayer}, this strategy is particularly compelling for foundational model development in QML.
In this vein, a universal and efficient framework has been proposed \cite{macaluso2023maqa} that emulates the outputs of various classical supervised algorithms through quantum computation. This framework is versatile, supporting a wide array of functions, and could act as the quantum analog to classical models that aggregate multiple and diverse functions.
From a computational perspective, this framework enables the creation of an exponentially large set of parametrized transformations of the input data. Importantly, the increase in the quantum circuit's depth that is required for this capability is only linear. Notable examples include the quantum versions of the classical single-layer perceptron \cite{macaluso2022variational,macaluso2020variational} and quantum ensembles \cite{https://doi.org/10.1049/qtc2.12087}.





\section{Current Focus in Quantum Machine Learning}\label{sec:current}

In the previous section, we explored various strategies for applying quantum computing to supervised learning tasks, which can be approached through multiple methodologies. This section outlines the current areas of focus in QML research, which include the challenges related to training these models and the complexity of establishing methodological metrics that can lead to quantum advantage.
Particular attention is given to QNNs and Quantum Kernels, which, given the current state of quantum technology, are the most promising for finding a potential advantage in the near term.

\subsection{Inductive Bias of Quantum Models}

In machine learning, inductive bias plays a crucial role in guiding algorithms towards effective generalization from training data to unseen instances \cite{baxter2000model,utgoff2012machine}. 
Usually, the inductive bias of a model is integrated into the functional form of $f(x;\theta)$ which aims to estimate the target variable of interest. For instance, the inductive bias of parametric models such as those based on linear regression is the linear relationship between the input features (or their augmented version) and the target variable. 
In SVMs, the assumption is that the data can be separated by a hyperplane in the feature space with as wide a margin as possible.
Other examples of inductive bias are for example local similarity rather than global structure in k-nearest neighbors or the layered structure of Neural Networks, which assume that data can be represented in a hierarchy of increasingly abstract features, allowing them to capture complex patterns and interactions among features. Each of these algorithms' inductive biases influences their approach to learning and their effectiveness in different problem settings, highlighting the importance of aligning the chosen model with the underlying data structure.

Recently, the exploration of inductive bias in QML models has been proposed as a promising avenue to achieve quantum advantage \cite{kubler2021inductive}. Similar to classical approaches, inductive bias in QML significantly influences the design and application of quantum algorithms for specific tasks. 
In the quantum domain, inductive bias can stem from three different sources. First, the method chosen for \emph{data embedding}, since, as explained in the previous sections, encoding classical data into quantum states corresponds to selecting a specific kernel \cite{schuld2021supervisedqml}. 
Second, the \emph{architecture of the quantum circuit} including the choice of quantum gates and their configuration, embodies certain presuppositions about the data distribution or the nature of the computational problem. This structural choice inherently biases the model towards specific solutions or interpretations of the data (as discussed in Section \ref{sec:classical-inspired}).
Third, the approach to measuring qubits, or the \emph{measurement strategy}, both during the training phase and inference, can significantly influence the model's inductive bias. Variations in measurement techniques can provide diverse insights into the quantum state, affecting the subsequent post-processing steps needed to derive the estimation \( f(x; \theta) \) \cite{schuld2018circuit}. These differences directly impact the learning dynamics and the algorithm's ability to generalize.

Importantly, indications of a potential quantum advantage via inductive bias are primarily found in scenarios where knowledge about the problem is integrated into quantum circuits \cite{kubler2021inductive}. 
Such advantages are more plausible when working with data generated by quantum processes, yet they tend to be less evident in classical datasets.

\subsection{Surrogates Models}

Training a quantum model using hybrid computation requires access to a quantum computer not only during the training phase but also for making predictions on new data. This requirement may be impractical for many real-world applications, considering that quantum computing resources will primarily be accessed through cloud-based services. Recently, the concept of a \emph{classical surrogate} or \emph{shadow models} \cite{jerbi2023shadows,schreiber2023classical} has been introduced. These are classical models derived from quantum models after training, capable of mimicking the quantum model's behavior.
This approach has been applied to a certain type of QML models based on the data-reuploading technique for data embedding, making them suitable for classical surrogation \cite{schreiber2023classical}. With a classical surrogate, the need for quantum hardware is limited to the training phase, highlighting the importance of achieving quantum advantages, only with respect to faster learning (e.g., the number of iterations in the the optimization) or better generalization.

However, the ability to create classical surrogates raises questions about the unique benefits of quantum methods. If classical surrogates can perform the same tasks, quantum models need to offer significant advantages in areas like generalization to justify their use. In this respect, recent works have shown that combining quantum models with classical shadow tomography techniques can tackle certain learning challenges that are beyond the reach of purely classical models, based on well-established cryptographic assumptions \cite{jerbi2023shadows}. This indicates that while classical surrogates can replicate many aspects of quantum models, there may still be unique advantages to quantum approaches, particularly for specific types of learning tasks.

\subsection{On Quantum Trainability}\label{sec:trainability}

The exploration of quantum machine learning algorithms as kernel methods or the implementation of surrogate models aims to determine the types of problems where near-term quantum computing can provide a practical advantage in machine learning. A fundamental assumption in these approaches is the efficiency in training quantum models, positing that it should be comparable to or better than that of classical models. However, this assumption encounters challenges specifically related to gradient-based optimization methods used in training quantum circuits.


\paragraph{Barren Plateau Problem}

The problem of barren plateau \cite{mcclean2018barren} emerges from the exponentially decreasing gradients of the cost function with respect to the parameters in quantum neural networks (QNNs). In the case of a single-qubit QNN, the cost function \( C(\theta) \) is influenced by the parameters \( \theta \), and the goal during training is to minimize this function. The gradient of \( C(\theta) \) with respect to \( \theta \) is represented as \( \nabla_\theta C(\theta) \).
The issue becomes pronounced when \( \nabla_{\theta} C(\theta) \) is negligibly small across a vast area of the parameter space. This is particularly critical in deep QNNs that contain multiple qubits and layers, leading to a substantial increase in the parameter count.

This phenomenon mirrors the vanishing gradient problem observed in classical neural networks, where the gradients of the cost function relative to the parameters diminish exponentially during backpropagation through several layers. Specifically, the vanishing gradient problem impedes the learning process as these gradients approach very small values.
Although both issues involve diminishing gradients, the barren plateau in QNNs is attributed to the intrinsic properties of quantum circuits and the optimization landscape's structure. In contrast, the vanishing gradient problem in classical neural networks mainly results from the activation functions' characteristics and the network's depth.

Addressing the barren plateau challenge requires specialized approaches tailored to quantum systems' unique properties \cite{kulshrestha2022beinit,cerezo2023does}. Furthermore, this issue raises significant concerns about the non-classical information processing capabilities of parameterized quantum circuits in barren plateau-free landscapes and the potential for superpolynomial advantages when executed on quantum hardware\cite{cerezo2023does}.

\paragraph{Lack of Quantum Backpropagation}

In classical deep learning, state-of-the-art models are trained using backpropagation, a method that efficiently computes gradients with computational and memory resources proportional to those required for calculating the function \(f\) that the neural network aims to approximate \cite{lecun2015deep,rumelhart1986learning,lecun1989handwritten}. Specifically, the time to calculate gradients for all parameters is at most the time to compute \(f\), scaled by a factor logarithmically related to the total number of parameters. This efficiency is pivotal for training large models, playing a significant role in the success of deep learning, especially in highly overparameterized settings \cite{allen2019learning}.

For quantum neural networks, the standard training approach on real quantum hardware involves the parameter shift rule to obtain gradients for parameterized quantum circuits \cite{crooks2019gradients}. Considering a quantum gate parameterized by a scalar \(\theta\), represented as \(U(\theta) = e^{-i \theta P}\), where \(P\) is the gate's generator (often a Pauli matrix), the parameter shift rule enables computing the gradient of an expectation value \(\langle \psi(\theta) | O | \psi(\theta) \rangle\), which is the classical output from a quantum circuit post-measurement, with respect to \(\theta\). This is formally given by:
\begin{equation}
\frac{\partial \langle O \rangle}{\partial \theta} = \frac{1}{2} \left( \langle O \rangle_{\theta + \frac{\pi}{2}} - \langle O \rangle_{\theta - \frac{\pi}{2}} \right).    
\end{equation}

Here, \(\langle O \rangle_{\theta \pm \frac{\pi}{2}}\) represents the expectation value of the observable \(O\) when the circuit is run with the parameter \(\theta\) adjusted by \(\pm \frac{\pi}{2}\).
Therefore, to estimate the gradient of a single parameter using this rule, the quantum circuit must be executed twice: once with the parameter incremented by \( +\frac{\pi}{2} \) and once with it decremented by \( -\frac{\pi}{2} \). For a quantum circuit with \( M \) parameters, a total of \( 2M \) executions are necessary to compute the gradients for all parameters. This computational cost associated with the parameter shift rule escalates rapidly for quantum circuits that have a large number of parameters, posing a significant challenge to the efficiency of training quantum models \cite{abbas2024quantum}. This problem is exacerbated by the barren plateau phenomenon, in which the magnitudes of gradients become increasingly small, further complicating the training process. Although recent studies suggest that improved scaling may be achievable through techniques such as shadow tomography \cite{abbas2024quantum}, these scaling challenges indicate that the overparameterization strategy, which has been effective in classical neural networks, might not be directly applicable to quantum models. This is primarily due to the exponential growth in computational and memory demands required for gradient calculations in extensive quantum systems.

\section{Promising research trajectories}\label{sec:primising}

The current trend in quantum machine learning research primarily revolves around theoretical exploration, utilizing classical machine learning and quantum information theory tools to dissect the intricacies of novel quantum models. This interdisciplinary method strives to deepen our grasp of quantum models from a learning theory perspective. However, relying solely on theoretical analysis might not suffice to unravel fundamental questions about how quantum computing could enhance learning algorithms.

Historically, machine learning advancements have largely stemmed from empirical methods, where training large-scale models provides valuable insights. This hands-on approach faces challenges in the quantum realm due to the nascent state of quantum technologies. Moreover, classical machine learning benefits from a robust theoretical foundation that aids in choosing specific models for particular problems, an aspect of sophistication QML is yet to achieve. 

Here we outline two distinct research directions that hold promise for advancing the field. Insights gained from quantum investigations in these areas could significantly enhance machine learning research and potentially unlock quantum advantages.

\subsection{Learn from (classical) experience}

The core principle of integrating quantum computing into supervised learning lies in leveraging computations that present exponential challenges in classical settings. This integration is facilitated by the relationship between quantum models and kernel methods, enabling the mapping of data into the high-dimensional Hilbert spaces that quantum computation provides. However, it is important to note that classical machine learning also utilizes kernel methods to implicitly operate within high- or infinite-dimensional function spaces \cite{shawe2004kernel}. Moreover, according to classical literature, neural networks almost systematically outperform kernel methods \cite{ghorbani2020neural}. Thus, identifying a quantum kernel that classical approaches find intractable does not necessarily mean that it will outperform neural networks. Consequently, focusing solely on the expressiveness of quantum machine learning models \cite{schuld2021effect,abbas2021power} does not fully address the potential scenarios where a quantum advantage could be achieved.

In this context, it may be beneficial to look beyond kernel methods and explore the fundamental components of the hypothesis classes that parametrized quantum circuits offer. For example, the foundational model for modern neural networks is the single-layer perceptron (SLP). Despite its simplicity compared to the complex architectures, the SLP model can be highly expressive. According to the universal approximation theorem \cite{hornik1989multilayer}, an SLP with a non-constant, bounded, and continuous activation function can approximate any continuous function within a closed and bounded subset, assuming a sufficient number of hidden neurons are available. However, despite this significant theoretical insight, SLPs are seldom used in practice due to the impracticality of managing large numbers of hidden neurons on classical devices.
Thus a relevant question in the quantum domain needs to be asked, which is:

\begin{center}
    \emph{Is there a quantum universal approximator capable of estimating any (classical) function?}    
\end{center}

Various efforts have been made in this direction; some attempt to adapt the SLP concept to quantum settings \cite{kapoor2016quantum,andrecut2002quantum,macaluso2022variational,macaluso2020variational}, while others employ PQCs to devise general methods for nonlinear approximations \cite{lubasch2020variational,10.1007/978-3-031-36030-5_14}. Yet, a definitive answer remains elusive. Nevertheless, identifying the \lq \lq quantum equivalent\rq \rq of the perceptron, which can act as a core model to build upon, might be a crucial stepping stone, just as it was for classical neural networks.

Another domain where QML practitioners might find valuable insights from classical literature is optimization. As highlighted in Section \ref{sec:trainability}, the efficiency of current quantum methods for gradient estimation falls short of classical backpropagation. Consequently, assessing the capabilities of a quantum model by directly comparing it to its classical counterpart, based solely on the number of parameters or any derived theoretical measure (e.g., effective dimension \cite{abbas2021power}), may not yield a fair evaluation. A more reasonable approach would be to compare quantum and classical models within specific contexts, acknowledging that classical models can be trained in an overparameterized regime, a luxury that quantum models do not match. This comparison could lead to more meaningful insights into the relative strengths and limitations of quantum models.

Furthermore, the issue of vanishing gradients did not deter the widespread adoption of neural networks; instead, it led to the development of various strategies to overcome this challenge. These strategies include the selection of activation functions that are less prone to vanishing gradients, careful weight initialization techniques, and the design of architectures specifically aimed at mitigating this issue. Therefore, rather than solely focusing on scenarios where the barren plateau phenomenon is absent \cite{cerezo2023does,pesah2021absence}, it may be more pragmatic to acknowledge that the problem of barren plateaus is likely to persist and to explore specific mitigation strategies.

The aforementioned approaches are grounded in a robust experimental approach, which seeks to identify specific (classical) settings that resonate with the classical machine learning community, as opposed to a purely methodological perspective.

\subsection{Building on the Shortcomings of Classical Methods}

Two of the main drawbacks of modern neural networks are their substantial demand for training data and the extensive computational resources needed for training. These models typically require large datasets to learn effectively, posing a considerable challenge in fields where data may be rare, sensitive, or costly to acquire. Additionally, the training process is resource-intensive, often necessitating millions of iterations to reach peak performance.
In light of these limitations, an intriguing question arises:
\begin{center}
    \emph{Can quantum models perform well in scenarios where labeled data is scarce, and a less data-intensive approach is preferred?}    
\end{center}

 Initial strides in this direction suggest promise. 
Recent studies have initiated exploration in this area \cite{liu2021rigorous,huang2021information}. One investigation \cite{liu2021rigorous}, capitalizing on the classical computational complexity associated with discrete logarithms has shown that with suitable feature mapping, a quantum computer could learn the target function with exponentially fewer data points than what would be feasible for any classical algorithm. Concurrently, another study \cite{Huang_2021} delved into generalization bounds and identified that the high expressiveness of quantum models could potentially hamper generalization. This insight led to the development of a heuristic aimed at optimizing dataset labels to enhance the learnability of the dataset by quantum computers, thereby challenging classical counterparts.
Also, evidence indicates that quantum models might achieve comparable generalization to classical models with far fewer training examples \cite{caro2022generalization}. This potential for efficiency in learning from limited data offers a compelling avenue for further exploration in quantum machine learning.  Thus, instead of generating synthetic data for testing quantum models, or using a small subset from large datasets, an alternative approach could be to compare the performance of classical and quantum models in scenarios characterized by a scarcity of annotated data. 

Another strategy might involve restricting the number of training iterations. For inference with large models, it could prove more practical to train a quantum model with a limited number of parameters and then derive classical surrogates from it. All these approaches should be underpinned by experimental evidence, which, despite being challenging to obtain, remains within the realm of feasibility.

Furthermore, there is a noticeable bias toward physical problems in quantum machine learning, stemming from its origins in quantum physics rather than as a discipline focused on learning. This bias is reflected in the numerous theoretical studies and the limited scope of experimental approaches. Although simulating quantum systems is challenging—which is a primary reason for the need for quantum computers \cite{feynman2018simulating}—current simulation tools can handle a few tens of qubits, allowing for training at a scale somewhat comparable to classical methods. Yet, the standard practice often involves using just a few qubits to demonstrate the viability of a proposed method rather than to showcase any potential advantage.

\section{Conclusion}\label{sec:conclusion}
In this paper, we have explored quantum machine learning through a classical lens. We began with a general overview of supervised learning problems, distinguishing between parametric and non-parametric methods. This distinction is crucial in the quantum context, leading to two distinct QML approaches. Parametric methods in quantum computing often involve convex optimization procedures that translate into solving linear systems of equations, where the Harrow-Hassidim-Lloyd (HHL) algorithm enables exponentially faster quantum routines. This approach, known as Fault-Tolerant Quantum Machine Learning (FT-QML), hinges on the successful execution of the HHL algorithm, requiring a fault-tolerant quantum computer. Its key characteristics include a single quantum computation for parameter estimation and learning capabilities that mirror classical methods.

With the advent of NISQ devices, a new approach has emerged based on hybrid quantum-classical computation. Here, a parametrized quantum circuit is treated as a black box, with parameters iteratively adjusted through training to minimize the cost function and enhance generalization. This method leverages the quantum computer for each optimization iteration, holding the promise of surpassing classical learning capabilities by exploiting QNNs.

We also highlighted the ongoing quest for a quantum advantage in QML research, amidst challenges in scaling optimization compared to classical neural networks. Discussions included the relationship between quantum models and kernel methods, and the use of surrogate models. Our aim was to provide a fresh perspective on QML research, advocating for a constructive dialogue with classical machine learning principles to address problems of mutual interest, and ensuring that advancements are beneficial in the near term.

In conclusion, contrary to some assertions in the QML literature, the current state of research does not definitively favor quantum models over classical ones. The true potential of quantum advantage may only be realized with the availability of fault-tolerant quantum computers. However, similar to the evolution of classical machine learning and neural networks, where methodological advancements often spur hardware developments, we believe a parallel trajectory is plausible for QML. It is reasonable to anticipate that the maturation of quantum machine learning will not only benefit from concurrent hardware improvements but may also influence the direction of these technological advancements.

\bibliographystyle{unsrt}      
\bibliography{references.bib}
\end{document}